%% file: main.tex
\newif\ifanonymous
\anonymousfalse

\documentclass[letterpaper, 10 pt, conference]{ieeeconf}  

\IEEEoverridecommandlockouts                              

\usepackage{times}
\usepackage{caption}
\usepackage{helvet}
\usepackage{courier}
\usepackage{graphicx}
\usepackage{booktabs}
\usepackage{makecell}
\usepackage{multirow}
\usepackage{rotating}
\usepackage{amsmath,amssymb,amsfonts}
\usepackage{tabularx}
\usepackage{siunitx}
\usepackage{textcomp}
\usepackage{xcolor}
\usepackage{dsfont}
\usepackage{comment}
\usepackage{algorithmic}
\usepackage{bbold} 
\usepackage{tikz}
\usepackage[table]{xcolor}  

\usepackage[pagebackref,breaklinks,colorlinks]{hyperref}
\usepackage[capitalize]{cleveref}

\newcommand\copyrighttext{%
  \footnotesize This work has been accepted for publication at the 2026 IEEE International Conference on Robotics and Automation (ICRA). \textcopyright 2026 IEEE. Personal use of this material is permitted. Permission from IEEE must be obtained for all other uses, in any current or future media, including reprinting/republishing this material for advertising or promotional purposes, creating new collective works, for resale or redistribution to servers or lists, or reuse of any copyrighted component of this work in other works.}

\newcommand\copyrightnotice{%
\begin{tikzpicture}[remember picture,overlay]
\node[anchor=south,yshift=10pt] at (current page.south) {\fbox{\parbox{\dimexpr\textwidth-\fboxsep-\fboxrule\relax}{\copyrighttext}}};
\end{tikzpicture}%
}

\crefname{figure}{Fig.}{Figs.}
\Crefname{figure}{Fig.}{Figs.}

\crefname{table}{Tab.}{Tabs.}
\Crefname{table}{Tab.}{Tabs.}

\crefname{equation}{Eq.}{Eqs.}
\Crefname{equation}{Eq.}{Eqs.}

\title{\LARGE \bf \LARGE \bf
DD-MDN: Human Trajectory Forecasting with Diffusion-Based Dual Mixture Density Networks and Uncertainty Self-Calibration}

\ifanonymous
    \author
    {
        Anonymous ICRA 2026 Submission
    }
\else
    \author
    {
        Manuel Hetzel\thanks{\textsuperscript{1}The authors are with the Faculty of Engineering, University of Applied Sciences Aschaffenburg, Germany. E-mail: \{manuel.hetzel,kerim.turacan,hannes.reichert,konrad.doll\}@th-ab.de.}\textsuperscript{1}
        \and
        Kerim Turacan\textsuperscript{1}
        \and
        Hannes Reichert\textsuperscript{1}
        \and
        Konrad Doll\textsuperscript{1}
        \and
        Bernhard Sick\thanks{\textsuperscript{2}The author is with the Intelligent Embedded Systems Lab, University of Kassel, Germany. E-mail: \{bsick\}@uni-kassel.de.}\textsuperscript{2}
    }
\fi

\ifanonymous
  \hypersetup{
    pdfauthor={Anonymous},
    pdftitle={Paper Under Double-Blind Review},
  }
\else
\fi

\ifanonymous
  \providecommand{\orcid}[1]{}%
  \providecommand{\email}[1]{}%
\fi
\begin{document}

\maketitle
\copyrightnotice
\thispagestyle{empty}
\pagestyle{empty}

\input{sections/0_abstract.tex}

\input{sections/1_intro}
\input{sections/2_sota}
\input{sections/3_methodology}
\input{sections/4_experiments}
\input{sections/5_conclusion}

\bibliographystyle{IEEEtran}
\bibliography{references}

\end{document}

%% file: sections/0_abstract.tex
%
\begin{abstract}
Human Trajectory Forecasting (HTF) predicts future human movements from past trajectories and environmental context, with applications in Autonomous Driving, Smart Surveillance, and Human-Robot Interaction. While prior work has focused on accuracy, social interaction modeling, and diversity, little attention has been paid to uncertainty modeling, calibration, and forecasts from short observation periods, which are crucial for downstream tasks such as path planning and collision avoidance. We propose DD-MDN, an end-to-end probabilistic HTF model that combines high positional accuracy, calibrated uncertainty, and robustness to short observations. Using a few-shot denoising diffusion backbone and a dual mixture density network, our method learns self-calibrated residence areas and probability-ranked anchor paths, from which diverse trajectory hypotheses are derived, without predefined anchors or endpoints. Experiments on the ETH/UCY, SDD, inD, and IMPTC datasets demonstrate state-of-the-art accuracy, robustness at short observation intervals, and reliable uncertainty modeling. The code is available at: \url{https://github.com/kav-institute/ddmdn}.
\end{abstract}

%% file: sections/1_intro.tex
\section{\large Introduction}
\label{sec1_introduction}
Human Trajectory Forecasting (HTF) is crucial for enabling intelligent autonomous systems such as autonomous driving~\cite{method_motiondiffuser,method_loki}, robot guidance~\cite{misc_rudenko_survey}, and service robots to operate safely among humans~\cite{method_robust_human_following}. It aims to estimate the most probable future states of human motion that exhibit intrinsic multimodality: identical histories can branch into diverse futures influenced by environmental and social constraints, as well as evolving intentions. Forecasting multiple diverse future trajectories is insufficient for downstream tasks, such as path planning and decision-making, which internally rely on probability estimates. Therefore, it is essential to model appropriate, calibrated uncertainty estimates for each discrete forecast. Numerous studies support the added value of probabilistic output modeling with uncertainty-related forecasts compared to discrete forecasts without any uncertainty-related output for path planning and situational assessment~\cite{method_bayesian, misc_av_challenges, misc_planning}.

Over the past few years, HTF research has gained significance, with several dozen publications emerging each year, all employing stochastic trajectory prediction as the primary forecasting objective. It aims for the best positional accuracy. A model outputs a distribution from which $K$ hypotheses are derived with at least one sample closely matching the Ground Truth (GT). However, these methods often lack evaluation of their underlying distributions, uncertainty quantification, and reliability calibration. Cognitive biases, such as the Dunning–Kruger effect, highlight misplaced confidence~\cite{misc_kruger}, which is mirrored by neural predictors that often over- or underestimate confidence~\cite {misc_calibration}, thereby undermining downstream decision-making processes. We demonstrate that accuracy-focused HTF methods are prone to incorrect confidence estimates due to their focus on K-shot accuracy. Moreover, the vast majority do not provide probability scores for the forecasted hypotheses. Both represent research gaps we aim to address.

We propose DD‑MDN (Diffusion‑based Dual Mixture Density Network): an end‑to‑end probabilistic HTF model unifying multimodal accuracy with self‑calibrated uncertainty from the very first observations. A few‑shot diffusion backbone and dual Mixture Density Network (MDN) generate uncertainty calibrated residence areas alongside confidence‑related anchor‑paths, ensuring aleatoric uncertainty modeling, probability‑sorted K‑shot hypotheses ranking, and robustness to short observation periods. Experiments on the ETH/UCY, SDD, inD, and IMPTC datasets demonstrate state-of-the-art (SOTA) accuracy, calibrated uncertainty estimates, and robustness, without requiring predefined waypoints or knowledge distillation.

%% file: sections/2_sota.tex
\section{\large Related Work}
\label{sec2_related_work}
Most HTF methods internally use probabilistic approaches and can therefore be described as probabilistic models. However, their results are deterministic (i.e., discrete future trajectories without uncertainty estimates); therefore, with respect to the final output, such methods can also be described as deterministic. The designation depends on the respective perspective. In the following, we refer to a model's output as either deterministic or probabilistic classification.
\subsection{Deterministic Methods}
Deterministic forecasting methods aim to generate at least one discrete forecast that is as close as possible to the GT.~\cite{htf_survey, survey_huang, survey_zhang} provide detailed surveys of deterministic HTF methods, including the latest developments and achievements. Social-\textbf{LSTM}~\cite{method_social_lstm} introduced social pooling for interactions, enhanced by Group-LSTM and SS-LSTM. \textbf{CNNs} capture spatial features and demonstrate ways to include map data (Y-net~\cite{method_y_net}, Next~\cite{method_next}). \textbf{GNNs} model agent interactions using nodes and edges (Social-STGCNN~\cite{method_social_stgcnn}, GroupNet~\cite{method_group_net}). \textbf{GANs} such as Social-GAN~\cite{method_social_gan} and its adaptations Sophie~\cite{method_sophie} combine encoded features with Gaussian noise and learn discriminators for trajectory validation. \textbf{CVAEs} explicitly learn conditional trajectory distributions (Trajectron++~\cite{method_trajectron++}, PECNet~\cite{method_pec_net}). \textbf{Normalizing Flow} networks (STGlow~\cite{method_stglow}, FlowChain~\cite{method_flowchain}) explicitly model data distributions via invertible transformations. Finally, \textbf{Transformer}-based architectures have recently excelled by incorporating self-attention and denoising diffusion, representing the current SOTA in deterministic HTF (LED~\cite{method_led}, SingularTrajectory~\cite{method_singular}, MoFlow~\cite{method_moflow}). All methods focus on social-interaction handling and positional-accuracy modeling, achieving continuous improvements in both.
\subsection{Probabilistic Methods and Uncertainty Modeling}
Uncertainty handling and the generation of accurate probabilistic forecasts are treated significantly less frequently in HTF than in deterministic approaches. ~\cite{htf_survey} provides an extensive overview of probabilistic and deterministic HTF methods. In addition, the reliability of the estimates and the calibration evaluation are rarely considered, leaving a research gap. Social-STAGE~\cite{method_social_stage} ranks multimodal Gaussian distributions using probabilities for each mode. TUTR~\cite{method_tutr} uses a Transformer architecture to forecast $K$ trajectories with associated probabilities encoded from their internal data distribution. In \textit{PSU-TF}~\cite{method_psu_tf}, uncertainties from the perception level are included in the forecasting process, and in~\cite{method_bayesian}, Bayesian Deep Learning techniques are used to provide realistic uncertainty estimates on the computed forecasts. All listed methods aim to partially or fully integrate uncertainty estimation, but none address uncertainty calibration or reliability evaluation. Some works, such as~\cite {method_mdn}, utilize MDNs to model continuous probability distributions with Confidence Levels (CL) and provide uncertainty evaluation, a methodology comparable to our contribution but without denoising diffusion, discrete hypothesis generation/evaluation, or context integration.
\subsection{Contributions} 
With DD-MDN, we address the gaps outlined above. Our main contributions are:
\begin{itemize}
  \item An end-to-end HTF framework coupling a few-shot denoising diffusion backbone with a dual MDN, producing uncertainty-calibrated residence areas and probability-ranked trajectory forecasts.
  \item Direct modeling of aleatoric uncertainty, yielding well-calibrated likelihood-ranked hypotheses without post-hoc recalibration.
  \item State-of-the-art accuracy, calibration, and short-observation robustness on various HTF datasets.
\end{itemize}

%% file: sections/3_methodology.tex
\section{\large Methodology}
\label{sec3_methodology}
\begin{figure*}[ht]
    \centering
    \includegraphics[width=\textwidth, trim=0 2 0 2, clip]{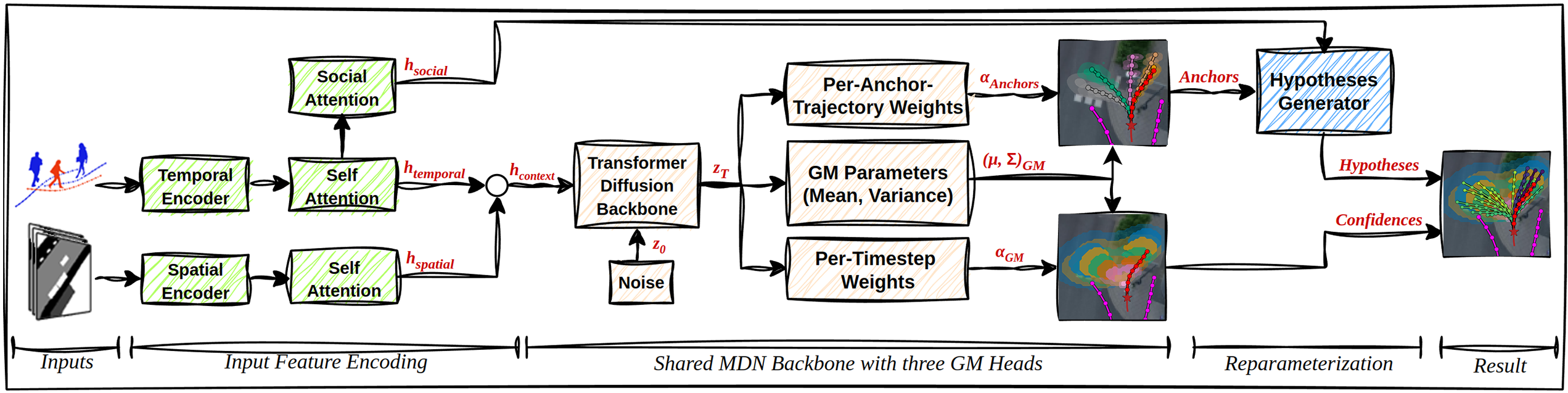}
    \caption{Architectural overview of DD-MDN. Input encoding blocks are green, probabilistic blocks are yellow, and deterministic ones are blue.}
    \label{fig:sec3_architecrture_overview}
    \vskip -6mm
\end{figure*}
We first define the HTF problem, then provide a high‑level overview of the architecture, and describe our input encoding. Next, we explain how DD-MDN performs probabilistic modeling with self-calibration, and dynamic mode pruning is implemented. Finally, we detail the semi-unsupervised denoising diffusion process and the generation of $K$ confidence-enriched discrete trajectory hypotheses.
%
%
\subsection{Problem Statement}\label{sec3_problem_statement}
HTF aims to forecast the future 2D positions of multiple agents based on their past movements and environmental context. Let $A$ be the number of agents, $T_{\mathrm{hist}}$ the history length, and $T_{\mathrm{fut}}$ the forecast horizon. We define:
\begin{equation}
    \mathbf{X}_a^{\mathrm{in}} = \bigl\{\mathbf{x}_{tp}^a \in \mathbb{R}^2 \;\bigm|\; tp = 1,\dots,T_{\mathrm{hist}}\bigr\}
  \label{eq:x_in}
\end{equation}
\begin{equation}
    \mathbf{Y}_a^{\mathrm{gt}} = \bigl\{\mathbf{x}_{tf}^a \in \mathbb{R}^2 \;\bigm|\; tf = T_{\mathrm{hist}}+1,\dots,T_{\mathrm{hist}}+T_{\mathrm{fut}}\bigr\}
  \label{eq:y_gt}
\end{equation}
\begin{equation}
    \mathcal{X}^{\mathrm{in}} = \bigl\{\mathbf{X}_a^{\mathrm{in}}\bigr\}_{a=1}^A \quad, \quad \mathcal{Y}^{\mathrm{gt}} = \bigl\{\mathbf{Y}_a^{\mathrm{gt}}\bigr\}_{a=1}^A
  \label{eq:xy_sets}
\end{equation}
The model takes as input all past trajectories, denoted as $\mathcal{X}^{\mathrm{in}}$, and an occupancy grid $\mathbf{G}$ that encodes scene context. To capture multimodality and aleatoric uncertainty, we predict for each agent a set of $K$ hypotheses $\dot{\mathcal{Y}}_a$ with corresponding probabilities $\dot{\mathcal{P}}_a$, see~\Cref{eq:y_hypos}, ensuring that at least one high-probability trajectory sample $\dot{\mathbf{Y}}_a^k$ per agent lies close to the GT.
\begin{equation}
  \dot{\mathcal{Y}}_a = \bigl\{\dot{\mathbf{Y}}_a^k \bigr\}_{k=1}^K
  \quad\text{with}\quad
  \dot{\mathcal{P}}_a = \bigl\{\dot{P}_a^k \bigr\}_{k=1}^K
  \label{eq:y_hypos}
\end{equation}
Explicitly, we learn a probabilistic model $F_{\Theta}\bigl(\dot{\mathcal{Y}}_a, \dot{\mathcal{P}}_a \mid \mathcal{X}^{\mathrm{in}},\,\mathbf{G}\bigr)$ with parameters $\Theta$, which jointly models multimodality and uncertainty. Because future paths are inherently uncertain, we adopt the standard SOTA stochastic trajectory prediction procedure and extend it as described in~\Cref {sec1_introduction} with calibrated aleatoric uncertainty estimates and CLs. 
%
%
\subsection{Architecture}\label{sec3_architecture}
\begin{figure}[ht]
    \centering
    \vskip -2mm
    \includegraphics[width=\linewidth, trim=0 0 0 0, clip]{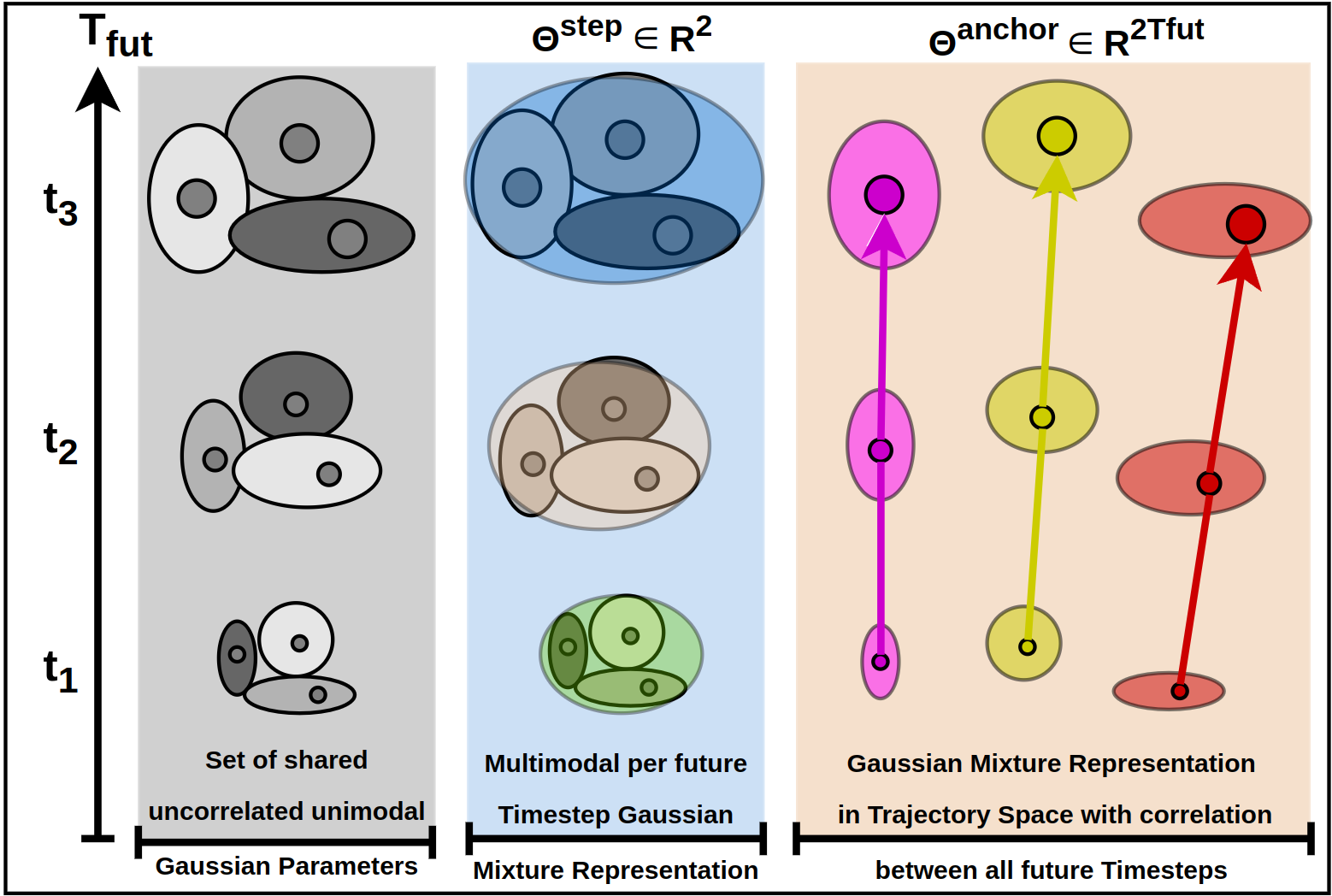}
    \caption{Schematic illustration of the two GM types, f.l.t.r.: \textbf{First:} Uncorrelated Gaussian parameters per-timestep (gray). \textbf{Second:} Representation in $\mathbb{R}^2$ space. \textbf{Third:} per-anchor-trajectory GM representation in $\mathbb{R}^{2T_{fut}}$ trajectory space.}
    \label{fig:sec3_mixture_overview}
    \vskip -6mm
\end{figure}

DD-MDN is an end-to-end learning framework comprising three parts: Encoding, Probabilistic Modeling, and Deterministic Hypotheses Generation. An architectural overview is illustrated by~\Cref{fig:sec3_architecrture_overview}. Classic encoder networks encode input data. A temporal encoder (LSTM) is used for temporal inputs (past agent motions), a spatial encoder (CNN) for spatial data (occupancy grid), and transformers (TF) for self- and social-attention matters, see~\Cref{sec3_input_encoding}. Probabilistic modeling processes temporal and spatial input features by using a dual MDN that comprises a shared TF-based denoising diffusion backbone and three probabilistic heads, which generate two distinct distributional representations. The architecture can handle various continuous probability distributions, including the Gaussian, Laplace, and Cauchy distributions. For this work, we use Gaussian Mixtures (GM). 
\begin{figure}[ht]
    \centering
    \includegraphics[width=\linewidth,, trim=0 0 0 0, clip]{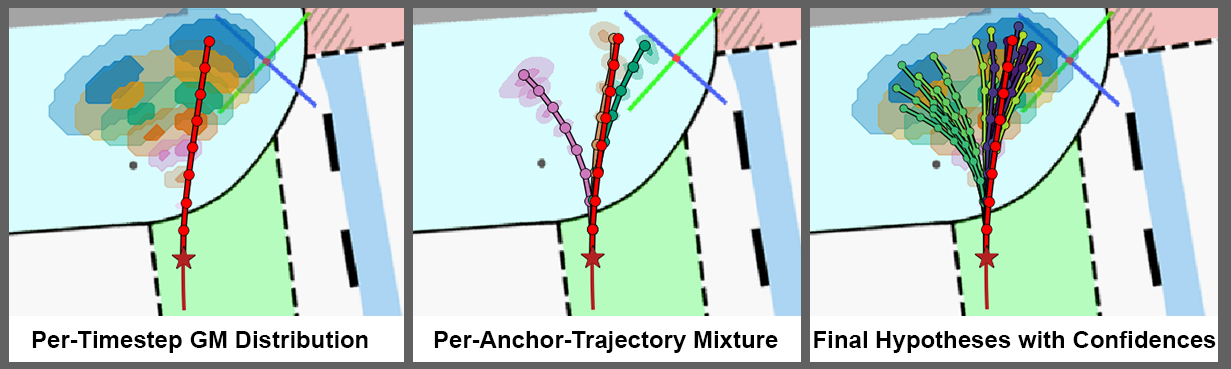}
    \caption{Internal GM representations, f.l.t.r.: \textbf{First:} Per-timestep GM representation in $\mathbb{R}^2$ space as 95- and 68 \% mixture CLs. \textbf{Second:} Per-anchor-trajectory GM representation in $\mathbb{R}^{2T_{fut}}$ trajectory space visualized by three anchor trajectories and their uncertainties. \textbf{Third:} Final discrete $K$ generated hypotheses. GT path is red.}
    \label{fig:sec3_mixture_explanation}
    \vskip -6mm
\end{figure}
We model future agent positions and their corresponding uncertainty estimates using two complementary GM representations, as illustrated in~\Cref {fig:sec3_mixture_explanation}. The first is a per-future-timestep GM $\Theta^{step}$, which provides calibrated one-step uncertainty at each future time step $tf$. For each timestep the conditional distribution of an arbitrary 2D position $\mathbf{x}_{tf}\in\mathbb{R}^2$ can be described as in~\Cref{eq:theta_step}:
\begin{equation}
 \Theta^{step}(\mathbf{x}_{tf})
  = \sum_{m=1}^{M} \alpha_{tf,m}
    \mathcal{N}\bigl(\mathbf{x}_{tf} \mid \boldsymbol\mu_{tf,m},\,\Sigma_{tf,m}\bigr)
  \label{eq:theta_step}
\end{equation}
We use multivariate normal distributions $\mathcal{N}$ with $\{\boldsymbol\mu_{tf,m}\in\mathbb{R}^2, \Sigma_{tf,m}\in\mathbb{R}^{2\times2}, \alpha_{tf,m}\}$ being the mean, covariance, and weight of mode $m$ at time $tf$. At each future time step, the model forecasts $M$ multivariate normal distributions and their corresponding weights, thereby constructing the GM $\Theta^{step}$. The second representation is called: per-anchor-trajectory GM $\Theta^{anchor}$, and it is used to generate multiple coherent future anchor trajectories $\mathbf{\widehat{Y}}$ from the same shared mean and covariance parameters ($\boldsymbol{\mu}_{tf,m}$, $\Sigma_{tf,m}$). Therefore, we form a joint GM in a $2T_{fut}$-dimensional trajectory space and define a set of $M$ future anchor trajectories as $\mathcal{\widehat{Y}}_a = \bigl[\mathbf{\widehat{Y}}_1,\mathbf{\widehat{Y}}_2, ...,\mathbf{\widehat{Y}}_M\bigr] \in\mathbb{R}^{2T_{fut}}$. For each mode $m$ in $\Theta^{anchor}$, we stack the shared means and covariances as described in~\Cref{eq:anchor_means} and~\Cref{eq:anchor_covs}.
\begin{equation}
  \widehat{\boldsymbol\mu}_{m}
  = \begin{bmatrix}
      \boldsymbol\mu_{1,m} \\[1pt]
      \vdots \\[1pt]
      \boldsymbol\mu_{T_{fut},m}
    \end{bmatrix}
  \in\mathbb{R}^{2T_{fut}}
  \label{eq:anchor_means}
\end{equation}
\begin{equation}
  \widehat{\Sigma}_{m} = \operatorname{blockdiag}\bigl(\Sigma_{1,m},\,\dots,\,\Sigma_{T_{fut},m}\bigr) \in\mathbb{R}^{2T_{fut}\times2T_{fut}}
  \label{eq:anchor_covs}
\end{equation}
We deliberately omit off-diagonal cross-time covariances to keep $\widehat{\Sigma}_m$ tractable; multiple mixture modes instead capture temporal diversity. Furthermore, we assign anchor trajectory weights $\widehat{\alpha}_{m}$ via a softmax over $m$. The resulting joint mixture over a position in the joint trajectory space $\widehat{\mathbf{x}}\in\mathbb{R}^{2T_{fut}}$ is:
\begin{equation}
  \Theta^{anchor}(\widehat{\mathbf{x}})
  = \sum_{m=1}^{M} \widehat{\alpha}_{m}
    \;\mathcal{N}\bigl(\widehat{\mathbf{x}}\mid \widehat{\boldsymbol\mu}_{m},\,\widehat{\Sigma}_{m}\bigr)
  \label{eq:theta_anchor}
\end{equation}
The per-anchor-trajectory distribution $\Theta^{anchor}$ is represented as a single multimodal GM in the trajectory space, where each mode $m$ represents a discrete future trajectory along with its associated variance. ~\Cref{fig:sec3_mixture_overview} illustrates the principle. $\Theta^{step}$ and $\Theta^{anchor}$ share the same core Gaussian parameters $\{\boldsymbol\mu_{tf,m},\,\Sigma_{tf,m}\}$ predicted by the shared backbone and core parameter head. Initially, the parameter pairs (mean and variance) are uncorrelated. This is established through the GM representations. The per-timestep GM and weights $\alpha_{tf,m}$ connect parameter pairs to form a related GM for each future timestep, a timestep-based clustering. However, there is no correlation between the individual time steps. Therefore, the means in this representation type cannot represent realistic future trajectories over time. That is where the per-anchor-trajectory GM representation and the $\mathbb{R}^{2T_{fut}}$ trajectory space come in. Here, the model is encouraged to make reasonable time-related connections among means across timesteps to generate realistic future trajectories. Those are used as anchor trajectories to derive the final $K$ hypotheses. This dual-GM design yields both trustworthy per-timestep uncertainty estimates and natural, time-consistent future anchor trajectories. ~\Cref{sec3_probabilistic_modeling} and~\Cref{sec3_unsupervised_generative_diffusion_learning} provide additional details. Finally, the Hypotheses Generator takes $\Theta^{step}$ and $\Theta^{anchor}$ and processes temporal and social input features to generate $K$ uncertainty-related discrete future trajectory hypotheses using affine reparameterization sampling, further described in~\Cref{sec3_deterministic_hypotheses_generation}.
%
%
\subsection{Input Encoding}\label{sec3_input_encoding}
The environmental context (map data, static objects, and vehicles) is represented as a single-layer occupancy grid centered on the target agent within a 10-meter radius. It distinguishes walkable areas (e.g., pedestrian walks) from inaccessible ones (e.g., buildings, walls). To align image coordinates $[u,v]$ with world coordinates $[x,y]$, we apply two Coordinate Convolutions (CordConv)~\cite{misc_cordconvs}, one per dimension, linking spatial and temporal features. A small CNN with self-attention processes the resulting three-layer grid into spatial features $\mathbf h_{spatial}$. Past trajectories of all pedestrian agents are encoded by an LSTM followed by self-attention, producing temporal features $\mathbf h_{temporal}$. Finally, Social-Out-Way Attention~\cite{method_social_out_way_attention}, an augmented Transformer with exit gates, filters irrelevant or distant neighbors without fixed thresholds. Each head outputs relevance and exit probabilities, pruning uninformative links end-to-end and yielding more accurate, socially compliant forecasts. Passing $\mathbf h_{temporal}$ through this module yields the social feature vector $\mathbf h_{social}$.
%
%
\subsection{Probabilistic Modeling}\label{sec3_probabilistic_modeling}
We aim for uncertainty self-calibration on a per-timestep basis to ensure that the probability density of $\Theta^{step}$ reflects the data's genuine aleatoric uncertainty so that we can derive trustworthy and reliable confidence areas, which describe how likely the target agent will reside within at each future timestep $tf$. Training with the Negative Log-Likelihood (NLL) corresponds to minimising the continuous logarithmic score, a strictly proper scoring rule~\cite{misc_gneiting_strictly}. Consequently, the network learns both the central tendency and the dispersion of the predictive distribution directly from the data, yielding an aleatorically calibrated forecast without the need for manual tuning of the covariance.

\textbf{Unimodal Case:} 
A neural network parameterizes mean and covariance as functions of input trajectory $\mathbf{X}^{in}$ and weights $W$. Minimizing the NLL aligns predicted distributions with GT data. For a single Gaussian, the NLL decomposes as
\begin{equation}
    \textit{NLL}_{tf} =
    \underbrace{\tfrac{1}{2}\, d_{tf}^2(\mathbf{x}_{tf}, W, \mathbf{X}^{in})}_{\text{error term}}
    + \underbrace{\tfrac{1}{2}\,\log \det \Sigma_{tf}(W, \mathbf{X}^{in})}_{\text{residual term}} .
\end{equation}
Here $d_{tf}^2$ is the Mahalanobis distance of $\mathbf{x}_{tf}$ to the predicted mean. The error term penalizes inaccurate means, while the residual term penalizes excessive covariance, reflecting the calibration–sharpness trade-off~\cite{misc_gneiting_probabilistic}.

\textbf{Multimodal Case:} 
Human motion is often multimodal, modeled with a Gaussian mixture:
\begin{equation}
    GM(\mathbf{x}_{tf}) = \sum_{m=1}^{M} \alpha_{tf,m}\;\mathcal{N}_{tf,m}(\mathbf{x}_{tf}), 
\end{equation}
\begin{equation}
    \textit{NLL}_{tf} = -\log (GM(\mathbf{x}_{tf})).
\end{equation}
Unlike the unimodal case, the NLL has no additive split. Components far from $\mathbf{x}_{tf}$ contribute little, resulting in gradients that pull the means $\boldsymbol{\mu}_{tf,m}$ toward the data. Covariances are implicitly regularized: inflating $\Sigma_{tf,m}$ lowers density at the GT point, so diffuse modes contribute negligibly and self-penalize. Mixture weights $\alpha_{tf,m}$ are normalized via softmax, concentrating on components that better match the data. During training, modes specialize, collectively capturing multimodality with calibrated probabilities.

\textbf{Mode Pruning:}
Not all $M$ modes are always necessary to represent reliable forecasts. We prune low-weight components via an epoch-dependent threshold $\delta(e)$, increasing from $\delta_0$ to $\delta_f$ over $E$ epochs:
\begin{equation}
  \delta(e) = \delta_0 + (\delta_f - \delta_0)\,\tfrac{e}{E}, \quad e=1,\dots,E.
\end{equation}
A sigmoid gate $G_m(e)$ filters each weight:
\begin{equation}
  G_m(e) = \sigma\!\Bigl(\tfrac{\alpha_m - \delta(e)}{\eta(e)}\Bigr), \quad
  \dot\alpha_m(e) = \tfrac{G_m(e)\,\alpha_m}{\sum_{j=1}^M G_j(e)\,\alpha_j}.
\end{equation}
Here $\eta(e)$ anneals from $\eta_0$ to $\eta_f$ for smooth soft-to-hard gating. Active modes are those with $\dot\alpha_m(e)>\delta(e)$, yielding $M^\star$ effective components whose weights renormalize to 1. This allows the model to adapt the mixture complexity to each input, retaining only the essential modes.
\subsection{Unsupervised Transformer Diffusion\label{sec3_unsupervised_generative_diffusion_learning}}
Denoising diffusion models learn the gradient of the log-density by corrupting observable GT data with a known Markov noise process and then training a neural network to invert that corruption. We transplant this idea into the parameter space of continuous probability distributions themselves, here using the shared multivariate normal parameters $\mathcal{N} \bigl(\mathbf{x}_{tf} \mid \boldsymbol{\mu}_{tf,m}, \,\Sigma_{tf,m}\bigr)$ and the separated weights $\alpha_{tf,m}$ and $\widehat{\alpha}_{m}$. The theoretical motivation for this parameter-space approach is to provide a generative prior over the manifold of valid distributions, ensuring global temporal coherence that standard point-estimate MDNs often lack. These parameters lie on a highly structured manifold, specifically the mean space $\mathbb{R}^{2}$, the simplex for weights, and the Symmetric Positive Definite (SPD) covariance cone. 

While direct Mean-Squared Error (MSE) regularization ignores this complex geometry, the iterative denoising process serves as a global regularizer that integrates local corrections with a context-aware score. Validity on the SPD manifold is maintained implicitly: the NLL objective serves as a geometric constraint, whereby degenerate or non-SPD covariance structures lead to a sharp drop in the probability density at the GT point. This triggers a self-penalizing gradient that steers the diffusion backbone back toward valid, stable regions of the parameter space. In our case, the observables are time-series data on future positions, whereas the continuous probability distributions are latent. Since no corresponding GT distribution for $\Theta^{step}$ and $\Theta^{anchor}$ exists, we cast them as the outputs of a conditional diffusion prior: starting from pure noise, the model is only guided by the NLL of the induced densities w.r.t. the GT trajectory $\mathbf{Y}^{gt}$ and a context feature $\mathbf h_{context}$ conditioned through Feature-wise Linear Modulation (FiLM)~\cite{misc_film}. For the shared diffusion backbone $f_{\boldsymbol\phi}$ , let $\mathbf z_{0}\sim\mathcal N(\mathbf 0,\mathbf I)$ be Gaussian noise in a learnable latent space. For $\tau=1,\dots,T_{Diff}$ diffusion steps, we iterate as described in~\Cref{eq:diffusion_step}.
\begin{equation}
    \mathbf z_{\tau}=f_{\boldsymbol\phi}\!\Bigl(\underbrace{\mathrm{FiLM}(\mathbf h_{context},\tau)}_{\text{context \& diffusion step}}, \mathbf z_{\tau-1}\Bigr)
  \label{eq:diffusion_step}
\end{equation}
$\mathbf h_{context}$ is the concatenation of $\mathbf h_{temporal}$ and $\mathbf h_{spatial}$. FiLM modulation injects this context at each diffusion step, so $f_{\boldsymbol\phi}$ learns a context-guided generative prior over valid GM parameters. Afterwards, the final latent state $\mathbf z_{T_{Diff}}$ branches into three heads to derive the dual mixture representations. Because both distribution representations are latent, optimization relies solely on the per-timestep and per-anchor-trajectory NLLs and their respective weightings (~\Cref{eq:diffusion_loss}).
\begin{equation}
\mathcal L_{prob}
        =\lambda_{step} NLL_{step}
        +\lambda_{anchor} NLL_{anchor}
  \label{eq:diffusion_loss}
\end{equation}
The right balance between $\lambda_{step}$ and $\lambda_{anchor}$ is crucial due to the shared core parameters ($\boldsymbol{\mu}_{tf,m}$, $\Sigma_{tf,m}$). $\lambda_{step}$ must be the main gradient driver; otherwise, per-timestep uncertainty calibration is not guaranteed, and on the other side, $\lambda_{anchor}$ must be in the right spot not to prevent per-timestep calibration while still having enough gradient influence to establish time-correlated and stable connections between means at different future timesteps to establish natural and realistic anchor-trajectories for the hypotheses generation. Across all experiments, $\lambda_{anchor} = 0.05$ and $\lambda_{step} = 1.0$ best meet the requirements, as evaluated through a parameter sweep. 

Back-propagating through the exact log-likelihoods drives the diffusion before generating mixture parameters that place high probability mass on the observed trajectory, both on a per-timestep basis and across the full trajectory, due to the dual mixture approach. This unsupervised, likelihood-based calibration signal enables the usage of the denoising diffusion process for GM parameter prediction without having corresponding GT mixture parameters. The process optimizes itself solely on the basis of future GT positions and their correlations. The parameter manifold is curved and coupled across timestep positions and complete trajectories. Iterative denoising regularizes this geometry: each step blends local corrections with a global, context-aware score, yielding (i) smooth evolution across the forecast horizon, (ii) diverse modes, and (iii) stable covariances. Ablations in~\Cref{sec_experiments} show that diffusion-augmented training improves the model's overall reliability compared to classic MDNs.
%
%
\subsection{Deterministic Hypotheses Generation}\label{sec3_deterministic_hypotheses_generation}
%
%
\textbf{Hypotheses Sampling:}
The generated anchor trajectories based on the correlated means of $\Theta^{anchor}$ are transformed into a finite set of $\mathcal{\dot{Y}}\!\in\!\mathbb R^{K\times 2T_{fut}}$ discrete trajectory hypotheses through an affine reparameterization sampling scheme. Each mode $m$ in $\Theta^{anchor}$ represents a time-stable mean-trajectory $\mathbf{\widehat{Y}}_{m}\!\in\!\mathbb R^{2T_{fut}}$ and a positive-definite block diagonal covariance matrix $\widehat{\Sigma}_{m}\!\in\!\mathbb R^{2T_{fut}\times2T_{fut}}$ together representing a future anchor trajectory and its variance. According to the anchor trajectory, the affine reparameterization is given by~\Cref {eq:differentiable_reparameterization}.
\begin{equation}
    \mathbf{\dot{\mathcal{Y}}}_{m} = \mathbf{\widehat{Y}}_{m} + \hat{\sigma}_{m} \mathbf{ \mathcal{B}}_{m}
\label{eq:differentiable_reparameterization}
\end{equation}
Here $\hat{\sigma}_{m}$ represents the overall variance scale of $\widehat{\Sigma}_{m}$ derived by the root-mean-square of the diagonal covariance values. In combination with a set of normalized residual trajectories of unit variance $\mathbf{\mathcal{B}}_{m}$, i.e., error patterns, we can derive a set of multiple future trajectory hypotheses $\mathcal{\dot{Y}}_{m}$ per anchor trajectory. The residual trajectories capture the typical ways a pedestrian might deviate from the central path. This two-step process cleanly separates the shape of likely deviations from their magnitude (mode-specific), yielding diverse, context-aware predictions while preserving full differentiability. A small MLP encoder takes the covariances as input and predicts a mode-specific scaling feature, $\mathbf{h}^{m}_{scale}$. This feature is concatenated with $\mathbf h_{temporal}$ and $\mathbf h_{social}$, passed through an MLP decoder, and simultaneously predicts the residuals $\mathcal B_{m}$. We take a mean field and a scalar variance scale that have already been learned from $\Theta^{anchor}_m$ and combine them with learning residual patterns guided by the input features. This affine reparameterization can be applied to each mode of $\Theta^{anchor}$, allowing for the generation of any number of hypotheses. 

Each active mode contributes at least its mean anchor trajectory. The remaining $K-M^{\star}$ hypotheses are apportioned in proportion to the pruned weights $\dot{\alpha}_m$ via a largest-remainder Hamilton rule quota $q_m$, ~\Cref{eq:hamilton}. Thus, the model learns for itself the number of hypotheses each anchor contributes to the final $K$ hypotheses.
\begin{equation}
    q_m = 1 + \bigl\lfloor r_m \bigr\rfloor,
    \quad
    r_m = \frac{\hat{\alpha}_m}{\sum_{j=1}^{M^{\star}}\hat{\alpha}_j}\,(K-M^{\star})
    \label{eq:hamilton}
\end{equation}
The leftover $L = K - \sum_mq_m$ slots are given one-by-one to the modes with the largest fractional remainders $r_m - \bigl\lfloor r_m \bigr\rfloor$. This guarantees $\sum_mq_m = K$ while maintaining a split as close as possible to the ideal proportional split. The number of contributed hypotheses depends on the model's learned confidence in specific modes, given the current situation.

\textbf{Hypotheses Loss:} To train the hypotheses generator, we combine three complementary objectives. Given $K$ forecasted trajectory hypotheses and the future GT trajectory, the generator loss is composed as described in~\Cref{eq:generator_loss}.
\begin{equation}
    \mathcal{L}_{\mathrm{hypo}} = \lambda_{MSE}\,\mathcal{L}_{MSE} +
    \lambda_{\mathrm{WTA}}\,\mathcal{L}_{\mathrm{WTA}} +
    \lambda_{\mathrm{Conf}}\,\mathcal{L}_{\mathrm{Conf}}
    \label{eq:generator_loss}
\end{equation}
The $\mathcal{L}_{MSE}$ term tends to pull the hypotheses toward the truth. In contrast, the Winner-Takes-It-All (WTA) $\mathcal{L}_{\mathrm{WTA}}$ term focuses on the most accurate candidate by softly emphasizing the best hypothesis with a temperature annealing schedule~\cite{misc_wta_for_traj}. The $\mathcal{L}_{\mathrm{Conf}}$ term keeps all hypotheses within a defined confidence region. For every future step $tf$ we compute a log‑probability threshold $\gamma_{tf}$ as the $(1-\beta)$-quantile of $\Theta^{step}_{tf}$. A hypothesis $k$ is penalised at that step only if its log‑probability falls below this threshold. We compute the worst deficit across the $K$ hypotheses and average it over batches and time. A hinge ensures that hypotheses lying inside the confidence region incur zero cost. At the same time, the maximization selects the single most offending hypothesis at each batch-timestep pair. This ensures that all hypotheses fall within the desired confidence interval.
%
%
\begin{figure*}[ht]
    \centering
    \includegraphics[width=0.95\textwidth, trim=0 2 0 2, clip]{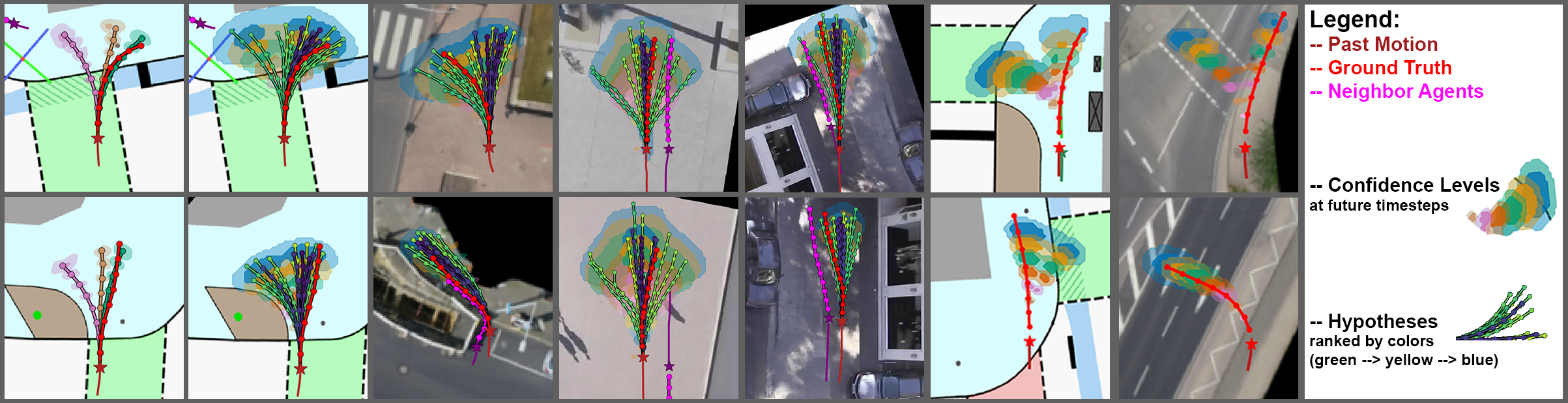}
    \caption{DD-MDN results in various datasets, f.l.t.r.: (1) IMPTC per-anchor-trajectory GM representation and (2) resulting final $K$ hypotheses with 68- and 95 \% CLs, (3-5) inD, SDD, and ETH examples, (6) IMPTC- and (7) inD per-timestep 68- and 95 \% isolated CL forecasts.}
    \label{fig:examples}
    \vskip -6mm
\end{figure*}

\textbf{Confidence estimation:}
We assign each forecasted hypothesis a confidence score $C_k\in[0,1]$ based on its likelihood under $\Theta^{step}$ using Monte Carlo (MC) percentiles. At each time $tf$, we draw $I$ MC samples $\{\mathbf{d}_{tf,i}\}_{i=1}^I \sim \Theta^{step}_{tf}$, compute their densities $\rho_{tf,i}$ as well as the hypothesis density $\rho_{tf,k}$ as described in~\Cref{eq:sample_and_hypo_densities}.
\begin{equation}
    \rho_{tf,i} = p\bigl(\mathbf{d}_{tf,i}\mid \Theta^{step}_{tf}\bigr) \;\;;\;\; \rho_{tf,k} = p\bigl(\mathbf{x}_{tf,k}\mid \Theta^{step}_{tf}\bigr)
\label{eq:sample_and_hypo_densities}
\end{equation}
\begin{equation}
      r_{tf,k} = \frac{1}{I}\sum_{i=1}^{I}\mathbf{1}\bigl[\rho_{tf,i}\ge \rho_{tf,k}\bigr] \in[0,1]
\label{eq:percentile}
\end{equation}
$r_{tf,k}$ is the fraction of MC samples whose density is at least as high as that of the hypothesis. We then estimate the upper-tail percentile of $\rho_{tf,k}$ among the $I$ samples (see~\Cref{eq:percentile}). To convert that percentile into a confidence weight, we quantize it into one of $J$ equal-width bins of size $ L = 1/J$. The bin index, ~\Cref{eq:steps}, identifies which percentile interval the hypothesis density falls into. We then assign a confidence weight equal to one minus the lower edge of that bin, so that hypotheses falling into lower percentiles (i.e.\ higher‐density samples) receive scores closer to 1. In contrast, outliers in higher percentile bins receive proportionally lower confidence, as shown in~\Cref{eq:conf}. Finally, the overall confidence of hypothesis $k$ is the average per‐timestep confidence.
\begin{equation}
    j_{tf,k} \;=\; \min\!\Bigl(\bigl\lfloor r_{tf,k}/L\bigr\rfloor,\;J-1\Bigr)
    \label{eq:steps}
\end{equation}
\begin{equation}
    c_{tf,k} \;=\; 1 \;-\; j_{tf,k} L
    \quad\in[0,1]
    \label{eq:conf}
\end{equation}
%
%
\subsection{Training Details}\label{sec3_implementation_details}
We train the model on an NVIDIA RTX 4090 and an AMD Ryzen 9950X using the AdamW optimizer, a cosine learning rate scheduler, and dynamic input-horizon scaling. The architecture is highly compact, comprising only \SI{4.5}{\mega\byte} of raw weights, representing a significant reduction in footprint compared with models such as LED (\SI{60.0}{\mega\byte}) or MoFlow (\SI{60.2}{\mega\byte}). Empirical analysis shows an average inference latency of \SI{15.5}{\milli\second} at $B=64$ and \SI{21.9}{\milli\second} at $B=128$. Under FP32 precision, the model maintains a low memory footprint of \SI{818.7}{\mega\byte} ($B=64$), including the fixed CUDA context. Using mixed-precision (FP16) or quantized (FP8) arithmetic can further reduce memory requirements by \SIrange{40}{60}{\percent}, thereby improving edge deployment suitability. All parameters and hyperparameters are available in the public repository. Because of the model's probabilistic nature, we report the average of three training runs.

%% file: sections/4_experiments.tex
\section{\large Experiments}
\label{sec_experiments}
\subsection{Experimental Setup}\label{sec3_experimental_setup}
We evaluate our framework on the ETH/UCY~\cite{dataset_eth}, SDD~\cite{dataset_sdd}, and inD~\cite{dataset_ind} benchmarks, providing pedestrian trajectories and scene context. We follow the ETH/UCY leave-one-out protocol~\cite{method_social_gan} and use the SDD/inD splits from~\cite{method_y_net}. All datasets utilize a \SI{2.5}{\hertz} sampling rate, with 8 observed frames (\SI{3.2}{\second}) and a 12-frame horizon (\SI{4.8}{\second}). Due to the limited size of standard benchmarks, we also utilize the full SDD, inD, and IMPTC~\cite{dataset_imptc_1,dataset_imptc_2} datasets ($\geq$\num{40000} samples), denoted as SDD\textsuperscript{*}, inD\textsuperscript{*}, and IMPTC\textsuperscript{*}, for robust reliability evaluation.

For accuracy, we report Best-of-$K$ Average and Final Displacement Errors ($min_{k}ADE$, $min_{k}FDE$) in meters $min_{k}ADE$ measures the mean $L_2$ distance of the best forecast $\mathbf{\dot{Y}}_k$ to the ground truth $\mathbf{Y}^{gt}$, while $min_{k}FDE$ evaluates only the endpoints. Forecast reliability is assessed via \textit{R$_{avg}$} and \textit{R$_{min}$}~\cite{method_mdn}, comparing predicted confidence levels $1-\beta(\mathbf{x})$ against observed frequencies $f_o(1-\beta)$. Scores are reported as percentages (100\% = perfect calibration) and visualized with Q–Q plots.
\subsection{Evaluation Results}\label{sec3_evaluation_results}
\begin{table}[htb]
    \setlength{\tabcolsep}{1mm}
    \centering
    \resizebox{\linewidth}{!}
    {%
        \begin{tabular}{c|ccccc|c}
            \toprule
            Method & ETH & HOTEL & UNIV & ZARA1 & ZARA2 & \textbf{AVG} $\downarrow$ \\
            \midrule
            \rowcolor{gray!20}  Trajectron                   & 0.61/1.03 & 0.20/0.28 & 0.30/0.55 & 0.24/0.41 & 0.18/0.32 & 0.31/0.52 \\
                                STAGE                        & 0.44/0.77 & 0.28/0.50 & 0.40/0.77 & 0.30/0.56 & 0.20/0.37 & 0.32/0.59 \\
            \rowcolor{gray!20}  TUTR                         & 0.40/0.61 & \textbf{0.11}/\underline{0.18} & \underline{0.23}/0.42 & 0.18/0.34 & 0.13/0.25 & 0.21/0.36 \\
                                LED                          & 0.39/0.58 & \textbf{0.11}/\textbf{0.17} & 0.26/0.43 & 0.18/ \textbf{0.26} & \underline{0.13}/ \underline{0.22} & 0.21/0.33 \\
            \rowcolor{gray!20}  SingTraj                     & 0.35/\underline{0.42} & 0.13/0.19 & 0.25/0.44 & 0.19/0.32 & 0.15/0.25 & 0.21/\underline{0.32} \\
                                MNRF                         & \textbf{0.26}/\textbf{0.37} & \textbf{0.11}/\textbf{0.17} & 0.28/0.49 & 0.17/0.30 & 0.14/0.25 & \textbf{0.19}/\underline{0.32} \\
            \rowcolor{gray!20}  MoFlow                       & 0.40/0.57 & \textbf{0.11}/\textbf{0.17} & \underline{0.23}/\underline{0.39} & \textbf{0.15}/\textbf{0.26} & \textbf{0.12}/\underline{0.22} & \underline{0.20}/\underline{0.32} \\
            \midrule               
                                \textbf{Ours}                & \underline{0.30}/0.46 & 0.13/\textbf{0.17} & \underline{0.23}/0.41 & \underline{0.16}/\underline{0.28} & \textbf{0.12}/\textbf{0.21} & \textbf{0.19}/\textbf{0.31} \\
            \bottomrule
        \end{tabular}
    }
    \caption{Stochastic evaluation on the \underline{ETH/UCY} benchmark with $min_{k}ADE/FDE$ (K=20) \underline{in meters}. \textbf{Bold} represents the best result, \underline{underline} represents the second best result.}
    \label{tab:sec4_eth_stochastic_eval}
    \vskip -2mm
\end{table}
We compare our method with the SOTA approaches in the established ETH/UCY, SDD, and inD benchmarks. Regarding ETH/UCY, the overall progress has reached a plateau since 2023; nonetheless, our model slightly outperforms the best available methods with a $min_{k}ADE/FDE$ of 0.19/0.31 m, as shown in~\Cref {tab:sec4_eth_stochastic_eval}. Concerning practical relevance, our method sets a new benchmark for momentary observation (shortened input horizon of two frames), outperforming the current SOTA SingTraj~\cite{method_singular} by over 20 \% (0.20/0.32 m vs. 0.25/0.40 m), see~\Cref {tab:sec4_eth_momentary_observation_eval}. The performance of most other methods drops significantly in this scenario. DD-MDN, on the other hand, is capable of generating highly accurate hypotheses with an input length of just two frames (position and velocity), which is very important for critical traffic situations with short observation and reaction times. 
\begin{table}[ht]
    \setlength{\tabcolsep}{1mm}
    \centering
    \fontsize{9pt}{9pt}\selectfont
    \begin{tabular}{lccc|c}
        \toprule
        Dataset & MID & EigenTraj & SingTraj & \textbf{Ours} \\
        \midrule
        \rowcolor{gray!20}  ETH          & 0.63/1.05 & 0.46/0.76 & \underline{0.45}/\underline{0.67} & \textbf{0.32}/\textbf{0.51} \\
                            HOTEL        & 0.29/0.49 & \underline{0.17}/\underline{0.28} & 0.18/0.29 & \textbf{0.13}/\textbf{0.18} \\
        \rowcolor{gray!20}  UNIV         & 0.30/0.56 & 0.25/0.44 & \textbf{0.24}/\underline{0.43} & \textbf{0.24}/\textbf{0.42} \\
                            ZARA1        & 0.30/0.56 & \underline{0.19}/0.35 & \underline{0.19}/\underline{0.33} & \textbf{0.17}/\textbf{0.28} \\
        \rowcolor{gray!20}  ZARA2        & 0.22/0.40 & \underline{0.15}/\underline{0.27} & 0.17/0.28 & \textbf{0.12}/\textbf{0.21} \\
        \midrule
        \textbf{AVG} $\downarrow$   & 0.35/0.61 & \underline{0.25}/0.42 & \underline{0.25}/\underline{0.40} & \textbf{0.20}/\textbf{0.32} \\
        \bottomrule
    \end{tabular}
    \caption{Momentary observation evaluation on \underline{ETH/UCY} benchmark with $min_{k}ADE/FDE$ (K=20) \underline{in meters}.}
    \label{tab:sec4_eth_momentary_observation_eval}
    \vskip -2mm
\end{table}
\newline
On the SDD benchmark in~\Cref{tab:sec4_sdd_ind_stochastic_eval}, using the full input horizon length of 8 past timesteps, marked as (@8), DD-MDN matches the performance of the best available methods regarding $min_{k}ADE$, but misses a top spot for $min_{k}FDE$. Regarding momentary observation, marked as (@2), no other method has reported results for SDD yet; nonetheless, our method confirms the strong performance already presented in the ETH/UCY momentary evaluation, still reporting the second-best $min_{k}ADE$ after just two input frames.
\newline
The inD benchmark is less frequently used for evaluation. Here, as listed on the right side in~\Cref{tab:sec4_sdd_ind_stochastic_eval}, DD-MDN significantly outperforms the SOTA, delivering a 46\% and 38\% improvement in $min_{k}ADE/FDE$ and confirming the excellent momentary observation performance. 
\newline
The robustness of DD-MDN with respect to variations in input length is based on our probabilistic modeling approach, combined with ego-coordinates and the use of a dynamic input trajectory length during training. This effectively quadruples the input training pattern, thereby enhancing the fine-tuning of the mixture parameters and improving the distinction between input data clusters.
\begin{table}[ht]
    \setlength{\tabcolsep}{2mm}
    \centering
    \fontsize{9pt}{9pt}\selectfont
    \begin{tabular}{c|c||c|c}
        \toprule
        \multicolumn{2}{c||}{\textbf{SDD (px)}} & \multicolumn{2}{c}{\textbf{inD (m)}} \\
        \cmidrule{1-2}\cmidrule{3-4}
        Method & ADE/FDE $\downarrow$ & Method & ADE/FDE $\downarrow$ \\
        \midrule
        SocialVAE                               & 8.88/14.81 & Y-Net          & 0.55/0.93 \\
        \rowcolor{gray!20} HighGraph            & 7.81/\underline{11.09} & AC-VRNN        & 0.42/0.80 \\
        LMTraj                                  & 7.80/\textbf{10.10} & Agentformer     & 0.57/0.87 \\
        \rowcolor{gray!20} MNRF                 & \textbf{7.20}/11.29 & Goar-SAR       & 0.44/0.70 \\
        MoFlow                                  & 7.50/11.96 & Di-Long        & 0.37/0.59 \\
        \midrule
        \rowcolor{gray!20} \textbf{Ours (@8)}   & \textbf{7.19}/11.82 & \textbf{Ours (@8)} & \textbf{0.20}/\textbf{0.36} \\
        \textbf{Ours (@2)}                      & \underline{7.42}/12.43 & \textbf{Ours (@2)} & \underline{0.21}/\underline{0.38} \\
        \bottomrule
    \end{tabular}
    \caption{Stochastic evaluation with $min_{k}ADE/FDE$ (K=20). \textbf{Left:} \underline{SDD} benchmark in pixels. \textbf{Right:} \underline{inD} benchmark in meters.}
    \label{tab:sec4_sdd_ind_stochastic_eval}
    \vskip -2mm
\end{table}

Despite delivering excellent positional accuracy, DD-MDN is capable of handling uncertainty by providing a probability score for each discrete forecast, representing the model's confidence. Due to the probabilistic modeling and the use of the NLL as part of the training loss, the model aims to achieve the best possible calibration of aleatoric uncertainty. In~\Cref{tab:sec4_reliability_eval_benchmarks} (A - Comparison), we evaluate DD-MDNs' uncertainty calibration performance regarding the \textit{Reliability} metric compared to other SOTA methods. For methods lacking native density estimation, we utilized Kernel Density Estimation (KDE) with \num{1e4} samples as an approximation. Our method's approach is beneficial, resulting in the highest \textit{Reliability} scores across all benchmarks; specifically, the \textit{Reliability} scores of DD-MDNs are the most accurate and trustworthy. In the upper row of~\Cref{fig:sec4_calibration_plots}, the Q-Q plots for the Zara01 subtest are visualized. The results for the other ETH/UCY subtests are similar. Accuracy-focused models tend toward overconfidence with overly tight distributions. While not reaching perfect calibration, DD-MDN outperforms current SOTA methods by a significant margin.
\begin{table}[ht]
  \setlength{\tabcolsep}{1mm}
  \centering
  \fontsize{9pt}{9pt}\selectfont
  \begin{tabular}{c|c|cc || c|c|cc}
    \toprule
    \multicolumn{4}{c||}{\textbf{A - Comparison}} & \multicolumn{4}{c}{\textbf{B - Ablation}} \\
    \cmidrule{1-4}\cmidrule{5-8}
    Set & Method & \textbf{R$_{avg}$} & \textbf{R$_{min}$} &
    Set & Method & \textbf{R$_{avg}$} & \textbf{R$_{min}$} \\
    \midrule
    \multirow{3}{*}{\scriptsize\begin{sideways}\textbf{ETH}\end{sideways}} 
      & LED          & 84.4 & \underline{71.1} &
    \multirow{3}{*}{\scriptsize\begin{sideways}\textbf{IMPTC\textsuperscript{*}}\end{sideways}}
      & Linear       & 97.9 & 92.9 \\
      & SingTraj & \underline{85.9} & 61.9 &
      & MLP          & 97.6 & 90.8 \\
      & \textbf{Ours}& \textbf{91.8} & \textbf{76.4} &
      & Diffusion    & \textbf{98.3} & \textbf{95.1} \\
    \midrule
    \multirow{3}{*}{\scriptsize\begin{sideways}\textbf{SDD}\end{sideways}}
      & MID          & 83.4 & 72.8 &
    \multirow{3}{*}{\scriptsize\begin{sideways}\textbf{inD\textsuperscript{*}}\end{sideways}}
      & Linear       & 95.7 & 91.7 \\
      & MoFlow       & \underline{87.6} & \underline{77.9} &
      & MLP          & 98.3 & 94.4 \\
      & \textbf{Ours}& \textbf{94.3} & \textbf{84.6} &
      & Diffusion    & \textbf{99.2} & \textbf{98.0} \\
    \midrule
    \multirow{3}{*}{\scriptsize\begin{sideways}\textbf{inD}\end{sideways}}
      & Agentformer  & 85.8 & 77.7 &
    \multirow{3}{*}{\scriptsize\begin{sideways}\textbf{SDD\textsuperscript{*}}\end{sideways}}
      & Linear       & 92.4 & 80.2 \\
      & Di-Long      & \underline{89.2} & \underline{84.3} &
      & MLP          & 92.1 & 83.3 \\
      & \textbf{Ours}& \textbf{98.3} & \textbf{94.8} &
      & Diffusion    & \textbf{93.6} & \textbf{83.8} \\
    \bottomrule
  \end{tabular}
  \caption{Uncertainty calibration evaluation with \textit{Reliability} (R$_{avg}$, R$_{min}$) \underline{in \%}. \textbf{Left (A)}: \underline{Benchmark} comparison with SOTA methods. \textbf{Right (B):} Ablations on \underline{full datasets}.}
  \label{tab:sec4_reliability_eval_benchmarks}
  \vskip -2mm
\end{table}
\subsection{Ablation Studies}\label{sec3_ablation_studies}
\begin{figure}[ht]
    \centering
    \includegraphics[width=\linewidth, trim=0 0 0 0, clip]{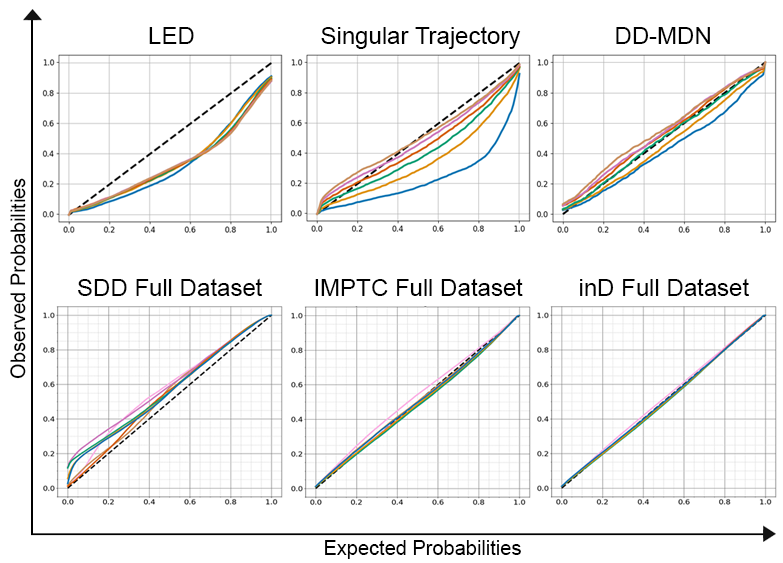}
    \caption{\textbf{Upper Row:} Calibration plots of SOTA methods (LED, SingularTraj) and DD-MDN for ETH/UCY Zara01 \underline{benchmark}. \textbf{Lower Row:} Calibration plots for the SDD\textsuperscript{*}, IMPTC\textsuperscript{*} and inD\textsuperscript{*} \underline{full datasets}. Colors represent future timesteps (0.8, 1.6, ..., 4.8 s), the black dotted diagonal represents perfect calibration.}
    \label{fig:sec4_calibration_plots}
    \vskip -6mm
\end{figure}

To better demonstrate our methods' uncertainty handling and self-calibration capabilities, we took the full SDD\textsuperscript{*}, inD\textsuperscript{*}, and IMPTC\textsuperscript{*} datasets and extracted train-eval-test splits, each with (\texttt{>=} 40 K samples). The right side of~\Cref {tab:sec4_reliability_eval_benchmarks} (B - Ablation) represents the results obtained using identical training parameters for all three complete datasets. Provided that sufficient training and test data are available, DD-MDN achieves near-perfect average- and strong minimum \textit{Reliability} scores for the inD\textsuperscript{*} and IMPTC\textsuperscript{*} datasets. The bottom row of~\Cref{fig:sec4_calibration_plots} provides the corresponding Q-Q calibration plots. The results for SDD\textsuperscript{*} show a slight decline, suggesting that training may have ended prematurely.
\newline
Moreover, in~\Cref{tab:sec4_reliability_eval_benchmarks} (B - Abalation), we also demonstrate the positive influence of the diffusion-based backbone. Compared to classic MDN modeling backbones (Linear Layers or MLPs), the diffusion backbone slightly improves the average and significantly boosts the minimum \textit{Reliability} scores across all three complete datasets, demonstrating the architecture's strong ability to calibrate uncertainty. ~\Cref{fig:examples} illustrates multiple forecasting examples with varying representation levels from all the datasets and benchmarks used.

%% file: sections/5_conclusion.tex
\section{\large Conclusion}
\label{sec_conclusion}
This work introduced a diffusion-based dual-MDN model for HTF. It predicts calibrated residence areas and diverse trajectories, offering trustworthy forecasts and uncertainty estimates for downstream tasks. DD-MDN achieves superior accuracy compared to SOTA models, with significant gains at short observation periods, and sets a benchmark for uncertainty calibration in HTF. We further showed how denoising diffusion enhances MDNs without mixture ground-truth data by leveraging an unsupervised likelihood-based calibration signal. Remaining limitations are evident in crowded scenes, where forecasts can be underconfident and broad. Future work will incorporate richer environmental contexts and agent forecasts to improve interaction modeling.